\def\year{2018}\relax
\documentclass[letterpaper]{article} %DO NOT CHANGE THIS
\usepackage{aaai18}  %Required
\usepackage{times}  %Required
\usepackage{helvet}  %Required
\usepackage{courier}  %Required
\usepackage{url}  %Required
\usepackage{graphicx}  %Required

\usepackage[utf8]{inputenc} % allow utf-8 input
\usepackage[T1]{fontenc}    % use 8-bit T1 fonts
\usepackage{booktabs}       % professional-quality tables
\usepackage{amsfonts}       % blackboard math symbols
\usepackage{nicefrac}       % compact symbols for 1/2, etc.
\usepackage{microtype}      % microtypography
\usepackage{subfigure}
\usepackage{caption}
\usepackage{mathtools}
\graphicspath{ {images/} }

\usepackage{algorithm2e}

\usepackage[textsize=tiny]{todonotes} %Todo notes
\usepackage{color}
\newcommand{\etal}{\textit{et al}. }

\begin{document}
\title{Efficient K-Shot Learning with \\Regularized Deep Networks}
%\if1
\author{ Donghyun Yoo$^1$,  Haoqi Fan$^2$, Vishnu Naresh Boddeti$^3$, Kris M. Kitani$^1$ \\
  $^1$The Robotics Institute, School of Computer Science, Carnegie Mellon University\\
  $^2$Facebook \\
  $^3$Michigan State University \\
  %Pittsburgh, PA 15213 \\
  \texttt{\{donghyun,kkitani\}@cs.cmu.edu}, \texttt{haoqifan@gmail.com}, \texttt{vishnu@msu.edu } \\
}
%\fi
\maketitle

\begin{abstract} 

Feature representations from pre-trained deep neural networks have been known to exhibit excellent generalization and utility across a variety of related tasks. Fine-tuning is by far the simplest and most widely used approach that seeks to exploit and adapt these feature representations to novel tasks with limited data. Despite the effectiveness of fine-tuning, it is often sub-optimal and requires very careful optimization to prevent severe over-fitting to small datasets. The problem of sub-optimality and over-fitting, is due in part to the large number of parameters ($\approx 10^{6}$) used in a typical deep convolutional neural network. To address these problems, we propose a simple yet effective regularization method for fine-tuning pre-trained deep networks for the task of $k$-shot learning. To prevent overfitting, our key strategy is to cluster the  model parameters while ensuring intra-cluster similarity and inter-cluster diversity of the parameters, effectively regularizing the dimensionality of the parameter search space. In particular, we identify groups of neurons within each layer of a deep network that share similar activation patterns. When the network is to be fine-tuned for a classification task using only $k$ examples, we propagate a single gradient to all of the neuron parameters that belong to the same group. The grouping of neurons is non-trivial as neuron activations depend on the distribution of the input data. To efficiently search for optimal groupings conditioned on the input data, we propose a reinforcement learning search strategy using recurrent networks to learn the optimal group assignments for each network layer. Experimental results show that our method can be easily applied to several popular convolutional neural networks and improve upon other state-of-the-art fine-tuning based k-shot learning strategies by more than $10\%$.

\end{abstract}

\section{Introduction}
\label{sec:intro}
% keyword: k-shot learning, fine tuning, regularized network, regularizing a pre-trained network
%Regularization using "clustering by looking at activation"

% Evolution of deep learning
% About Deep Learning
% Deep learning has been expanding its use in various fields such as computer vision, speech recognition, text recognition. It has been achieved a lot of successes in these areas. 
% In particular, deep learning has shown excellent performance in computer vision tasks like object classification, object detection, object tracking, etc. Most of the superior methods that perform well in any competition are based on deep learning. Many successful network architectures have been proposed for vision tasks such as VGG network by \cite{vgg}, Inception by \cite{googlenet} and ResNet by \cite{residual}. 

% Deep learning shows excellent performance in various fields
% Recent networks are getting better and better performance.
% But a deep neural network requires big data because it has many parameters.
% A lot of data
% Fine tuning, domain transfer

% problem definition
Even as deep neural networks continue to exhibit excellent performance on large scale data, they suffer from severe over-fitting under learning with very low sample complexity. The growing complexity and size of these networks, the main factor that contributes to their effectiveness in learning from large scale data, is also the reason for their failure to generalize from limited data. Learning from very few training samples or $k$-shot learning, is an important learning paradigm that is widely believed to be how humans learn new concepts as discussed in \cite{thorpe1996speed} and \cite{li2002rapid}. However, $k$-shot learning still remains a key-challenge in machine learning. 

Fine-tuning methods seek to overcome this limitation by leveraging networks that have been pre-trained on large scale data. Starting from such networks and carefully adapting their parameters have enabled deep neural networks to still be effective for learning from few samples. This procedure affords a few advantages: (1) enables us to exploit good feature representations learned from large scale data, (2) a very efficient process, often involving only a few quick iterations over the small scale, (3) scales linearly to a large number of $k$-shot learning tasks, and (4) is applicable to any existing pre-trained networks without the need for searching for optimal architectures or training from scratch. 

\begin{figure}[t]
  \centering
  \includegraphics[width=8cm]{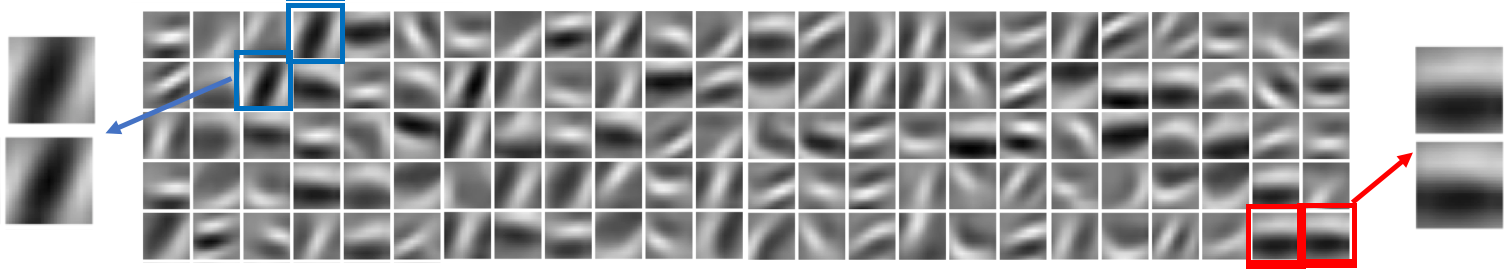}
  \caption{Convolutional Filters are often highly correlated. First layer visualization of LeNet. Correlated filters highlighted in red and blue.}
  \label{fig:conv-filters}
\end{figure}

Unfortunately, fine-tuning can be unstable especially when the amount of training data is small. Large deep neural networks typically are comprised of many redundant parameters, with the parameters within each layer being highly correlation with each other. For instance consider the filters, shown in Fig.\ref{fig:conv-filters}, in the first layer of LeNet \cite{yann} that was learned on the MNIST dataset. A number of filters are similar to other filters, \emph{i.e.}, these filters functionally play the same role and tend to produce similar activations. The presence of a large number of correlated filters can potentially lead to over-fitting, especially when learning under a small sample regime. %Our key insight to alleviate this redundancy problem is (1) to group filters whose activations are correlated with each other, and (2) to update them jointly with a shared gradient. 

%It requires very careful optimization and can still result in over-fitting. We conjecture that this is due to the fact that fine-tuning approaches directly optimize all the model parameters using a single supervisory signal, typically, the classification loss on the $k$-shot dataset. This procedure allows all model parameters to be updated without any control, often resulting in over-fitting to limited data.

To stabilize the fine-tuning process, we propose a simple yet effective procedure to regularize fine-tuning based $k$-shot learning approaches. The key idea of our approach is to identify the redundancies in the parameters and constrain their updates during fine-tuning. This is achieved by first clustering the parameters in each layer of the network into multiple groups based on the similarity of their activations on a specific $k$-shot learning task. The parameters in each group share a common update while ensuring intra-group similarity and inter-group diversity of activations. By grouping the model parameters and guiding the fine-tuning process with more supervisory signals, our approach is able to reduce the capacity of the network, to mitigate over-fitting and improve the effectiveness of pre-trained networks for $k$-shot learning.

We make the following contributions in this paper, (1) \textbf{ grouping neuron by activations for layer-wise clustering of parameters} while enforcing intra-group similarity and inter-group orthogonality of group activations, (2) \textbf{a hybrid loss function} for $k$-shot learning consisting of cross-entropy loss as well as triplet loss among the $k$-shot data, the later providing more supervision for optimizing the model, and (3) \textbf{a reinforcement learning based mechanism to efficiently search for the optimal clustering of the parameters} across all the layers of the model. Our proposed $k$-shot learning approach affords the following advantages: (1) task agnostic approach to $k$-shot learning that does not rely on any task-specific prior knowledge, (2) is applicable to any network without having to change the original network structure, and (3) a general purpose technique for decomposing the parameter space of high capacity deep neural networks.

To demonstrate the effectiveness of our approach we experimentally evaluate it across two tasks: an one-shot domain-adaption task for matching images across three different domains and a $k$-shot transfer learning task. Our experimental results show that the proposed approach yields significant performance improvements over task agnostic fine-tuning approaches for small sample learning without the need for any task specific prior knowledge.

\section{Related Work}
\label{sec:relatedwork}

\noindent\textbf{$k$-shot Learning:} One of the earliest work on one-shot learning for object categories was proposed by Fei-Fei \etal \cite{feifei}. The authors developed a Bayesian learning framework with the premise that previously learned classes can inform a prior on the model parameters for a new class. %Subsequently, many different approaches have been developed along these lines.
Among recent work, powerful generative models have been developed that compose characters from a dictionary of parts \cite{2015iccvoneshot} or strokes \cite{lake}. Such generative models have shown great promise on datasets with limited intra-class variation. Siamese networks \cite{siamese} has been used to automatically learn feature representations where objects of the same class are closer together. Santoro \etal \cite{mann} proposed the memory-augmented neural networks with an external content based memory. Wang and Hebert \cite{Wang,wang2016learning} propose a regression approach from classifiers trained on small datasets to classifiers trained on large datasets. Vinyals \etal \cite{matchingnet} proposed matching networks that learns a non-parameteric $k$-nearest neighbor classifier through end-to-end learning, with the weights for the nearest neighbors are provided by an LSTM.  Ravi and Larochelle \cite{ravi2016optimization} proposed LSTM-based meta-learner that uses its state to represent the learning updates of the parameters of a classifier for $k$-shot learning. 
Hariharan and Girshick \cite{hariharan2016low} suggest a  novel squared gradient magnitude regularization technique and techniques to hallucinate additional training examples for small data classes.
While these approaches have state-of-the-art performance on $k$-shot learning problems, they often utilize specific architectures designed for these problems. In contrast, we explore a more general method that can reuse existing networks, by fine-tuning them for $k$-shot learning.

\noindent\textbf{Domain Adaptation:} These methods seek to adapt a pre-trained model trained on one domain (source domain) to another domain (target domain). \cite{Daume} proposed an adaptation method through feature augmentation, creating feature vectors with a source component, a target component, and a shared component. A Support Vector Machine (SVM) is then trained on this augmented feature vector. \cite{judy} used the feature representation of a pre-trained network like AlexNet that was trained on the 2012 ImageNet 1000-way classification dataset (\cite{imagenet}). The authors replace the source domain classification layer with a domain-adaptive classification layer that takes the activations of one of the existing network’s layers as input features. We are also interested in adapting a model learned on large scale data from the source domain to a model for the target domain with few examples. However, unlike these approaches, we propose a \textit{task adaptive regularization} approach that improves the adaptability of exiting pre-trained networks to new target domains with limited training samples.

\section{Proposed Approach}
\label{sec:approach}

Our focus in this paper is the task of $k$-shot learning by fine-tuning an existing pre-trained network. We consider the setting where the pre-trained network was learned on a source domain with large amounts of data and the $k$-shot target domain consists of very few samples. To avoid the pitfalls of overfitting when training with only a few examples, we propose the following strategy. (1) We first search for similar activations to identify redundant filters, and then group them in a source domain. (2) After identifing the redundant parameters, the pre-trained network is fine-tuned with group-wise backpropagaion in a target domain to regularize the network. The proposed (1) layer-wise grouping method and (2) model fine-tune by group-wise backpropagation effectively make the fine-tuning on $k$-shot samples more stable. However, our proposed grouping method has a significant hyper-parameter, the number of groups. Deciding the number of groups for each layer is a non-trivial task as the optimal number of groups may be different in each layer. (3) We suggest a hyper-parameter search method based on reinforcement learning to explore the optimal group numbers.

We now describe the three sub-components involved in our approach: (1) \textbf{grouping neurons by activations}, (2) \textbf{model fine-tuning for $k$-shot learning} and (3) \textbf{a reinforcement learning based policy for searching over the optimal grouping of the parameters}.

%\textcolor{blue}{(KK: This paragraph needs to be written in a more argumentative style, making more of an effort to connect the three components logically. Right now it is just a list of components without any explicit logical connection.)} 

\subsection{Grouping Neurons by Activations (GNA)}
%\vspace{1mm}
%\noindent\textbf{Grouping by Activation.} 

\begin{figure}[tb]
  \centering
  \includegraphics[height=2.0cm]{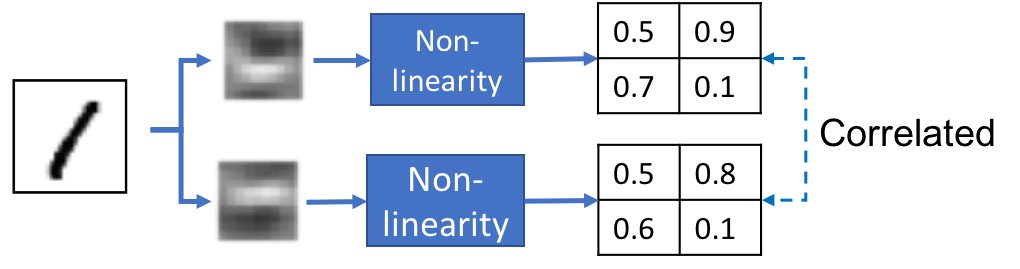}
  \caption{Correlated filters in activation point of view }
  \label{fig:similar activation}
\end{figure}

To identify redundant parameters to be grouped together for more stable fine-tuning, we define correlated filters as filters which have similar activations conditioned on a set of training images. We would like to group these correlated filters as a means of regularizing the network. Fig.~\ref{fig:similar activation} illustrates a toy example of two convolutional filters with very correlated activations (heatmaps). Since the two filters have similar patterns, their outputs are very similar.

%\textcolor{blue}{
%Considering a simple general neural network, when there is a fully connect layer as shown in Fig~\ref{fig:correlated-orthogonal filters}, batch-type learning data is input and sent to layer $L$ through layer $L-1$ to calculate the output of activation function of layer $L$. In this figure, $A_i$ is the output of the non-linear activation function of each neuron $i$ in layer $L$. At this time, if we look at $A_i$, the activation output $A_i$ by the redundant parameters becomes similar to each other. $A_1$ and $A_2$ have similar outputs and can be said to be correlated. On the other hand, in the case of $A_1$ and $A_3$, the activation output is very different and can be said to be orthogonal.}

\begin{figure}[tb]
  \centering
  \includegraphics[height=5.0cm]{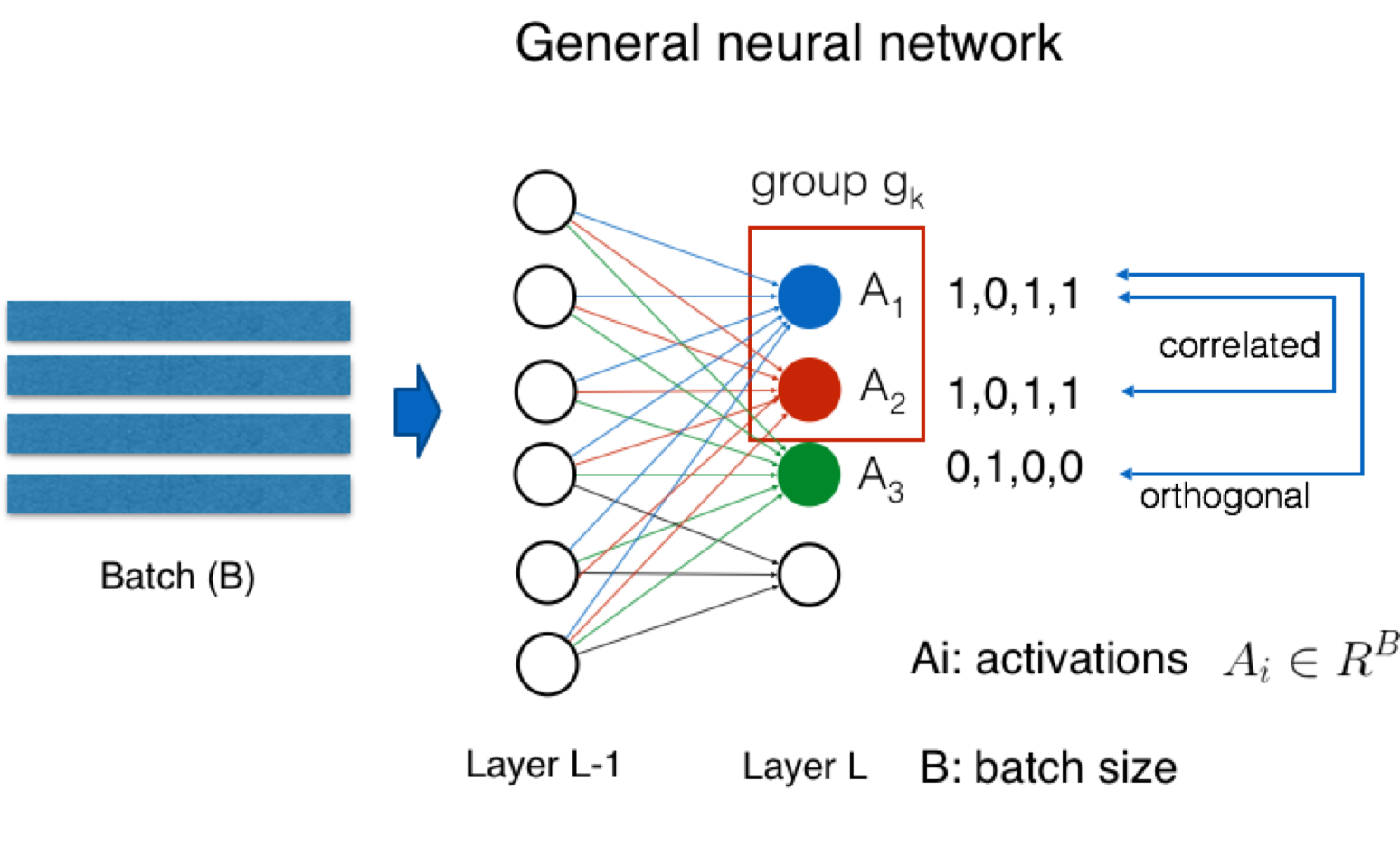}
  \caption{Correlation of parameters. Circles are neuron of a network. $A_i$ is activation of neuron $i$. $A_1$ and $A_2$ have similar actions, so they are correlated. Thus, we can group $A_1$ and $A_2$ to one group $g_k$.}
  \label{fig:correlated-orthogonal filters}
\end{figure}

Now consider the fully connect layer of a neural network illustrated in Fig~\ref{fig:correlated-orthogonal filters}. Given a batch of data $B$ as input, we can pass each data element (image) through the network to compute the activations at layer $L$. $A_i$ is the output of the non-linear activation function of the $i$-th neuron in layer $L$. If we compare activation $A_i$ to another activation $A_j$ over the input data, we can measure the correlation between neurons. In our example, $A_1$ and $A_2$ have similar output patterns over the batch image data whereas, $A_1$ and $A_3$ have different output patterns. This implies that $A_1$ and $A_2$ are good candidates for grouping.

In our proposed approach, we use a clustering algorithm to group similar neurons based on their activations over the  $k$-shot training data (\emph{e.g.}, one image for each category). In particular, we use $k$-means clustering to group the neurons and the number of clusters $k$ for each layer is learned via a reinforcement learning procedure described later. 

%\todo{make sure all of the k's are $k$ or not has variables.}
%\todo{use emph for all of the i.e., and e.g.,s}

%During training, the neurons associated to the same group will receive the same (average) gradient during back-propagation described next.

%\textcolor{blue}{
%The correlated parameters (i.e., parameters with similar activation output) can be grouped into a group. This grouping by activation outputs can effectively regularize deep networks with an average gradient propagation method instead of updating them individually. We group the filters by performing k-mean clustering on the activations of the respective filters. The grouping procedure is performed over the entire set of training images.}

%\begin{center}
%\label{average gradient}
%$\bigtriangledown P_a = \frac{1}{N} \sum_{i \in g_k} \bigtriangledown P_i $
%\end{center}
%where $P_i$ is a parameter in a group $g_k$, and $N$ is the number of parameters %in $g_k$. 

\subsection{Backpropagation with Groups}

%The presence of a large number of redundant filters can potentially lead to over-fitting, especially when learning under a small sample regime. 
Once the redundant parameter groups in each layer are identified, an effective regularization method is required during fine-tuning to prevent over-fitting. To restrain the redundant parameters overfiting, we can consider updating the parameters in a group with one gradient because the gradients of the redundant weights in the same group would be expected to be very similar to each other. 
%\todo{abrupt logical transition here. talking about gradients don't have anything to do with the last sentence. please fix. also check grammar carefully.} 
%The gradients of the redundant weights that are in (a) the same group would be expected to be very similar to each other since the parameters in each group are highly correlated and their activations have a similar effect on the subsequent layers \todo{try not to put more than one idea in a sentence, until you gain more experience writing. Split into two sentences}. 
From this insight, we update the parameters of each group $g_k$ by a shared gradient $W_a$ during learning to regularize the network. The shared update is computed as the average gradient of all the filters in the group \emph{i.e.},
%\begin{equation}
\label{eq:average_gradient_update}
$\Delta W_a = \frac{1}{|g_k|} \sum_{i \in g_k} \Delta W_i$,
%\end{equation}
\noindent where $\Delta W_i$ is the gradient of $W_i$.
We demonstrate the feasibility of this backpropagation by an average gradient with domain adaptation and transfer learning experiments described later.

\subsection{Loss Functions}

%$K$-shot learning presents a unique challenge to the learning paradigm 

%\todo{This first statement is meaningless. It has no direct connection to the design of the loss function. The first sentence should start with a motivating/justification of the importance of the loss function.}. 

The low sample complexity of typical $k$-shot learning results in extremely noisy gradient updates for each $k$-shot entity. To provide more supervisory signals to the learning process, we introduce a triplet loss to the network optimization objective. The triplet loss is similar to the one introduced by Schroff \etal \cite{facenet}.
% intuition of the triplet loss
The triplet loss serves the twin purposes of providing more supervisory signals to increase the separation between the $k$-shot entities as well as to reduce the noise in the gradient signal by averaging over larger number of loss terms %\todo{This should be mentioned earlier}.

%\todo{try to make your sentences shorter and split as necessary. This will help to avoid confusing sentences until your writing style gets better} 
%\todo{add the intuition of the triplet loss here. How is it different from any other alternative??? This will help the reader to understand why this is necessary. What happens if you just use the softmax classification loss???} 

We define the triplet loss for the $k$-shot learning problem as:
%\todo{math is a part of the sentence. You need a comma or semi-colon here}

\begin{equation}
\label{eq:tripletloss}
L_{triplet} = \sum_{i,j,k} \left[d\left(f(x_i),f(x_j)\right) - d\left(f(x_i), f(x_k)\right) + \alpha\right]_{+} 
\end{equation}%\todo{you need a comma here}
,where $f(x)$ is the output of the network for input $x$, $i$,$j$ are indices of samples belonging to the same class and $k$ is the index of sample belonging to a different class, $d\left(f(x_i),f(x_j)\right)$ is the distance between $f(x_i)$ and $f(x_j)$ and $[\cdot]_{+}$ denotes the margin maximizing loss. The distance can be the Euclidean distance $\|\cdot\|_2^2$ and total variation $\|\cdot\|_1$ for regression and classification tasks respectively.  We note that the triplet loss reduces to a margin loss for one-shot learning. The margin loss is defined as: %\todo{punctuation is missing!}

\begin{equation}
\label{eq:marginloss}
L_{margin}=\sum_{i,k} \left[- d\left(f(x_i), f(x_k)\right) + \alpha\right]_{+}.
\end{equation} %\todo{period is missing!!! please check all of your math equations for the proper punctuation.}

In addition to the classification loss described above, 
it is important to ensure that the intra-group activations are similar to each other, while the inter-group activations are orthogonal to each other. We augment the $k$-shot learning loss function with these two criterion during training. Let the activation of a filter $i$ in the $l$-th layer be $A_i^l$. The intra-group similarity loss is defined as: % \todo{:}

\begin{equation}
\label{ea:intraloss}    
L_{intra} = \sum_l\sum_k \sum_{i,j \in g_k} \left \| A^l_i - A^l_j \right \|_2.
\end{equation}

The inter-group orthogonality loss is defined as: % \todo{:}
\begin{equation}
\label{eq:interloss}    
{L}_{inter} = \sum_l\sum_{i, j} \| M^{l^\top}_i M^l_j \|^{2}_F 
\end{equation}, %\todo{comma}
\noindent where $M^l_i$ and $M^l_j$ are matrices with the activations of all the weights in group $g_i$ and $g_j$ respectively at the $l$-th layer and $\| \cdot \|^2_F$ is the squared Frobenius norm.

Our $k$-shot learning task is trained to optimize a combination of the loss functions described above. The total loss is described as the following equation:
\begin{equation}
\label{eq:totalloss}
L = L_{class} + \alpha L_{intra} + \beta L_{inter} +\gamma L_{triplet}
\end{equation}
\noindent ,where $\alpha$, $\beta$ and $\gamma$ are hyper-parameters that control the importance of each of the loss terms.

% 3-5 RL clustering
\subsection{Hyper-Parameter Search Through Reinforcement Learning}

The performance of the proposed approach is critically dependent on the number of clusters that the weights in each layer are grouped into. Manually selecting the number of clusters can lead to sub-optimal performance while an exhaustive search is prohibitively expensive. This problem is exacerbated as the number of layers in the network increases. Common methods for determining hyper parameters are brute force search, grid search, or random search. While brute force search is guaranteed to find the optimal solution, it is very time consuming and is usually intractable. Grid search is the most used method for hyper-parameter selection, but is still limited by the granularity of the search and can potentially end up being computationally expensive. On the other hand, surprisingly, Bergstra and Bengio \cite{Bergstra:2012:RSH:2188385.2188395} suggests that random search is more effective than grid search. Recently, Hansen \cite{dqn-hyperparam} proposed a reinforcement learning approach for determining hyper-parameters. Building upon this, Zoph and Le \cite{DBLP:journals/corr/ZophL16} proposed a neural network architecture to find the optimal hyper-parameter of a neural architecture through reinforcement learning. In this work, we adopt a similar approach to determine the optimal number of clusters in each layer of the network for $k$-shot learning.

\begin{figure}[t]
    \centering
    \subfigure[Hyper-Parameter Search Framework]{\includegraphics[width=0.35\textwidth]{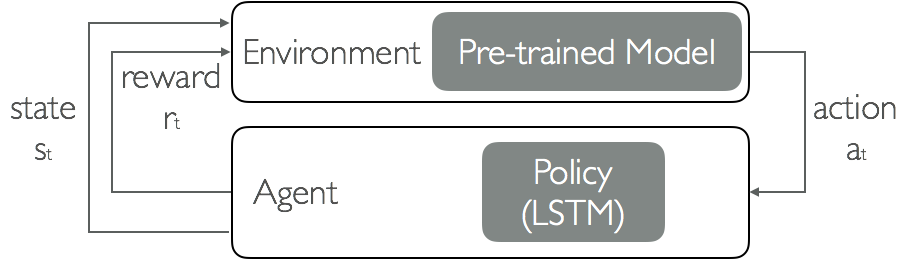}}
    \subfigure[Hyper-Parameter Policy Network]{\includegraphics[width=0.3\textwidth]{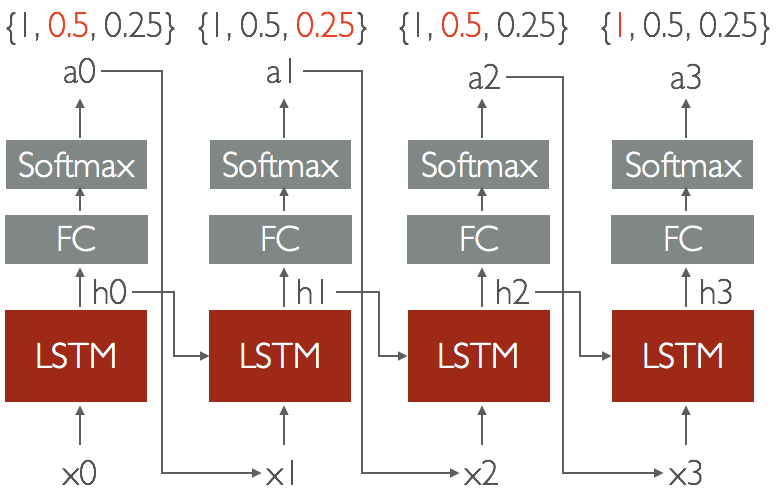}}   
    \caption{We adopt a deep reinforcement learning framework to search over the clustering hyper-parameters, number of groups. Furthermore, we adopt an LSTM as our policy network to determine the numbers of clusters in each layer.}
    \label{fig:lstm}
\end{figure}

We pose the hyper-parameter search problem as a reinforcement learning problem to find a locally optimal layer-wise group size for the entire network. Figure \ref{fig:lstm}(a) shows our reinforcement learning problem, where the environment is a pre-trained network that we wish to fine-tune for $k$-shot learning. Intuitively the policy network implicitly learns the relation between different groupings of the layer weights and the performance of the network. We model the problem as a fixed horizon episodic reinforcement learning problem where all actions (layer-wise prediction of number of clusters) have an equal affect on the final outcome. We represent the sequence of actions as $a_{1:L}$, where $a_l$ is the action at the $l$-th layer, predicting the number of clusters in the $l$-th layer. We define the state as a vector that has the number of groups of each layer.
\begin{equation}
    \label{ea:rl_state}
    S=\{n_1, n_2, ..., n_i\}
\end{equation}
,where $n_i$ is the number of groups in layer $i$.

Our agent's policy network is a Long Short-Term Memory (LSTM) by Hochreier and Schmidhuber \cite{hochreiter1997long} as shown in Fig. \ref{fig:lstm}(b) and is learned through a policy gradient method. The time horizon of the LSTM is equal to the number of layers in the pre-trained network. The output of the LSTM consists of a fully connected layer followed by a softmax layer to predict the probabilities of the action $a_l$ at the $l$-th layer. The input of the policy network at the $l$-th layer, $I \in \mathbb{R}^{N_a+1}$ is a vector created by the concatenation of the number of filters (a single integer) in the $l$-th layer and the output action at the previous layer (one hot encoding), where $N_a$ is the number of actions.  

We adopt the policy gradient method \cite{DBLP:journals/corr/ZophL16} to learn our agent's policy that maximizes the expected accuracy of the proposed fine-tuning process through parameter clustering since the cumulative reward $R$ is non-differentiable. We define agent's reward returned by the environment as the accuracy $A_{ft}$ of the fine-tuned model on the validation set for a valid action and -1 for an invalid action (impossible group size).

\begin{center}
\label{reinforce_rule}
\[R =
  \begin{cases}
    A_{ft} & \quad \text{if an action is valid} \\
    -1  & \quad \text{otherwise}\\
  \end{cases}
\]
\end{center}
$A_{ft}$ is the accuracy of the fine-tuned network of which parameters are clustered and calculated on the validation set. %\todo{a bit confusing}.  
In each episode the agent predicts a list of actions $a_{1:L}$ corresponding to the number of groups in the $L$ layers of the network. The parameters in each layer of the pre-network are clustered into the number of groups as determined by the action. The pre-trained network is then fine-tuned on the $k$-shot data until convergence, after which the validation accuracy of the network is recorded to use as a reward $R$ for the agent. The agent's policy network is then updated by backpropagating the gradients computed from the loss Eq. \ref{eq:rl_optimize}. As the episodes are repeated, the policy network's predictions inch closer to the optimal number of parameter clusters in each layer, in turn resulting in a gradual increase in the accuracy of the fine-tuning process.

To estimate the optimal clustering of the network's parameters, the policy network's parameters $\theta_c$ are optimized to maximize the expected reward J ($\theta_c$), computed over all future episodes from current state. %We need to maximize this expected reward instead of a current reward in one current state because the expected reward is the estimation of total reward that the agent receives when one episode finishes. 
\begin{equation}
\label{exp_reward}
J(\theta_c) = E_{P( a_{1:T} ; \theta_c )} [R]
\end{equation}

Since the reward signal is non-differentiable, we use an approximate policy gradient method to iteratively update the policy network. In this work, we use the REINFORCE rule from \cite{Williams} 
\begin{equation}
\label{eq:reinforce_rule}
\bigtriangledown _{\theta_c} J(\theta_c) = \sum_{t=1}^{T} E_{P( a_{1:T} ; \theta_c )} [\bigtriangledown _{\theta_c} log P(a_t | a_{(t-1):1};\theta_c)R ]
\end{equation}

The above quantity can be empirically approximated as:
\begin{equation}
\label{eq:rl_optimize}
\frac{1}{m} \sum_{k=1}^{m} \sum_{t=1}^{T}  \bigtriangledown _{\theta_c} log P(a_t | a_{(t-1):1};\theta_c) R_k
\end{equation}
,where $R_k$ is a reward of k episode, and $m$ is the number of episodes. $P(a_t|a_{(t-1):1};\theta_c)$ denotes the probability of a history of actions $a_{1:T}$ given policy-defining weights $\theta_c$. Our complete algorithm is presented in Algorithm \ref{algorithm1}

\begin{algorithm}[!h]
\SetAlgoLined
\caption{Grouping and average gradient update algorithm}
\label{algorithm1}
%\toprule
 Given a pre-trained network $M$, source domain samples and $k$ samples of target domain\\
 \hspace{1mm}
 
 \textbf{Step 1)} Grouping and fine-tuning in a source domain\\
 \For{each iteration}{ 
     Generate actions that change the numbers of groups for each layer by a recurrent policy network \\
     
     Set the numbers of groups for each layers \\
     
     Cluster the parameters of each layer in network M to $K$ groups by k-mean clustering\\
     
     Fine-tune the network M with source domain training samples \\

     $R_k$ $\leftarrow$ the validation accuracy of fine-tuned network \\
     
     Update recurrent policy network by policy gradient update equation Eq.  \ref{eq:rl_optimize}
    }
\hspace{1mm}
\textbf{Step 2)} Fine-tuning in a target domain\\
\For{each iteration}{
    \For{each group $g_i$, $i \in [1, K]$}{
    compute $L_{intra} = \sum_l\sum_k \sum_{i,j \in g_k} \left \| A^l_i - A^l_j \right \|_2 $ \\
    compute ${L}_{inter} = \sum_l\sum_{i, j} \| M^{l_i^\top} M^l_j \|^{2}_F $ \\
    compute $L_{triplet} = \sum_{i,j,k} \left[d\left(f(x_i),f(x_j)\right) - d\left(f(x_i), f(x_k)\right) + \alpha\right]$ \\
    }
    
    $L = L_{class} + \alpha L_{intra} + \beta L_{inter} +\gamma L_{triplet}$
    
    Update by average gradient
    $\Delta W_a = \frac{1}{|g_k|} \sum_{i \in g_k} \Delta W_i $ \\
}
%\bottomrule
\end{algorithm}

% It is a neural network architecture that has a number of hyperparameters. It has a number of different architectures. The validation accuracy is that the k-th neural network architecture achieves after being trained on a training dataset is $R_k$.

% To demonstrate the efficacy of learning such a policy network we evaluate its performance on a simulated toy problem. We consider a pre-trained network with six layers, and assume that each layer of the network consists of 10 filters. The optimal number of filters in each layer is set to $O=[2,2,2,5,7,7]$. For the purpose of the toy example the accuracy of the network is defined as $1-tanh (\| O - C \|)$, where $C$ is current value. The agent's action space, $\{ 0.125, 0.25, 0.5, 0.75, 1\}$, is defined as a multiplication factor that is applied to the number of filters in each layer to obtain the optimal number of filters in each group. The results of the training process are shown in Fig. \ref{fig:rl-simulation}. A brute force search under these settings would require $5^6 = 15626$ trials while the proposed reinforcement learning based search is able to determine the optimal clustering in about 400 trials.

% \begin{figure}[h]
%   \centering
%   \includegraphics[height=5cm]{fig-rl-simulation.eps}
%   \caption{Accuracy of simulation environment in hyper-parameter search by reinforcement learning}
%   \label{fig:rl-simulation}
% \end{figure}

\section{Experiment}
\label{sec:experiment}

The usefulness of the proposed method is verified through experiments on two tasks, domain adaptation and transfer learning. In both tasks, we show how our approach can be used to learn a new model from only a few number of examples. We present results of multiple baseline variants of our proposed approach, (1) \textbf{Fine-Tuning:} the standard approach of updating a pre-trained network on $k$-shot data with cross-entropy loss, (2) \textbf{Fine-Tuning+Triplet Loss:} updating a pre-trained network on $k$-shot data with cross-entropy loss and the triplet loss, (3) \textbf{GNA:} proposed orthogonal grouping of parameters with cross-entropy loss and manual hyperparameter search, (4) \textbf{GNA+Triplet Loss:} proposed orthogonal grouping of parameters with cross-entropy and triplet loss and manual hyperparameter search, (5) \textbf{GNA+Triplet Loss+Greedy:} proposed orthogonal grouping of parameters with cross-entropy and triplet loss and greedy hyperparameter selection, and (6) \textbf{GNA+Triplet Loss+RL:} proposed orthogonal grouping  of parameters with cross-entropy and triplet loss and RL based hyperparameter search.

\subsection{Domain Adaptation}

%\begin{figure}[tb]
%    \centering
%    \includegraphics[width=0.4\textwidth]{office-source.png} \\(a) %Amazon
%    \includegraphics[width=0.4\textwidth]{office-target.png} \\(b) %Webcam
%    \caption{Office dataset for domain adaptation}
%    \label{fig:office_dataset}
%\end{figure}

For this task, we consider the Office dataset introduced by \cite{saenko2010adapting} consisting of a collection of images from three distinct domains: Amazon, DSLR and Webcam. The dataset consists of 31 objects that are commonly encountered in office settings, such as keyboards, file cabinets, laptops \textit{etc}. We follow the experimental protocol used in \cite{judy}, and consider domain adaptation between the Amazon (source) and the Webcam (target) images. The experiment is conducted on 16 out of the 31 objects that are also present in the ImageNet dataset. Our pre-trained network is the ResNet-18 architecture by \cite{he2016identity} trained on the ImageNet dataset. Our action space for this experiment is the number of possible clusters in each layer $a = \{2^0,2^1,2^2,2^3,2^4,2^5,2^6,2^7,2^8\}$, or equivalently the action space is the number of possible groups per layer. We set the action space to $\{1, 2, 4, … , N_f\}$, where $N_f$ is the number of filters. The minimum number of groups is one. The maximum number of groups is the same as the number of weights. In this work, we define the actions (number of possible clusters) as $\{2^0, 2^1, ..., N_f\}$ to reduce the size of the action space and speed up the search process. However, in general, the action space can also be continuous like $\{1,2,3,....N_f\}$. 
%We consider this continuous action space for the future work.

\begin{figure}[tb]
    \centering
    \includegraphics[width=0.35\textwidth]{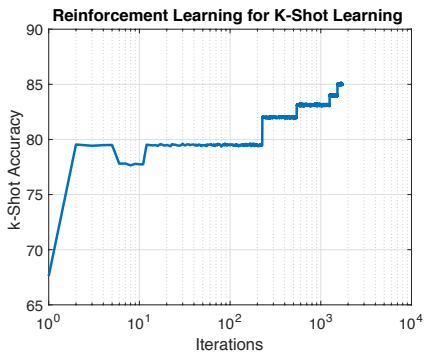}
    \caption{Progression of hyper-parameter search through reinforcement learning through the iterations for $K$-shot domain adaptation on Office dataset}
    \label{fig::rl_office}
\end{figure}

We use 20 source images per each class for clustering the parameters and fine-tune the model with 16 one-shot examples, one image per class. The performance of our proposed approach is compared to the baselines in Table \ref{results:office}, and Figure \ref{fig::rl_office} shows the progression of the reinforcement learning based hyperparameter search on the $k$-shot learning accuracy. 
\textit{Late fusion}~\cite{judy} and \textit{Daume}~\cite{Daume} are compared as the baselines. The \textit{Late fusion} and \textit{Daume} 
%\todo{use a different font or typeface to distinguish baselines from regular text}
use DeCAF-7 features in their works, but we also apply their method with ResNet-18 features for fair comparison with our method. For fine-tuning, the learning-rate is 0.01, and it is changed to 0.001 after 1000 iteration. We tried 10 random runs with randomly selected different dataset to get average performance.

We note that the clustering hyper-parameter search through the reinforcement learning is able to efficiently search the hyper-parameter space and find better parameter groupings compared to both manual and greedy search. For the \textit{manual baseline}, 
%\todo{need some kind of codeword here to show that this is a baseline}
we initialize the number of groups in all the layers to two and compute the accuracy of the network. We then compute the accuracy of the of the network by doubling and halving the number of groups in a layer. The action (doubling or halving) that results in higher accuracy is selected. We repeat this process and update the number of groups iteratively through the process described above.

For the \textit{greedy baseline(Greedy)}, 
%\todo{same comment here. use a different font to show that this is a baseline. It should match with the font used in the table and graphs}
we set the number of groups in the first layer to two and compute the accuracy of the original network. If the accuracy is greater than before, then the number of groups is doubled, otherwise we set the number of groups to the previous number and move to the next layer. We repeat this procedure until the last layer.

\begin{table}[tb]
\caption{Experimental Evaluation: k-shot domain adaptation on Office dataset}
\scalebox{0.85}{
\centering
\begin{tabular}{lll}
    \toprule
    \cmidrule{1-3}
    Method & Feature type & Accuracy(\%) \\
    \midrule
    \textit{Late fusion} \cite{judy} & DeCAF-7     & 64.29       \\
    \textit{Late fusion} \cite{judy} & ResNet-18   & 71.08       \\
    \textit{Daume} \cite{Daume}      & DeCAF-7     & 72.09        \\
    \textit{Daume} \cite{Daume}      & ResNet-18   & 76.25        \\
    Fine-Tuning               & ResNet-18   & 70.07              \\
    Fine-Tuning + margin loss & ResNet-18   & 70.34     \\
    \textbf{GNA}              & \textbf{ResNet-18}   & \textbf{79.94}         \\
    \textbf{GNA + margin loss}& \textbf{ResNet-18}   & \textbf{82.16}           \\
    \textbf{GNA + margin loss+Greedy}  & \textbf{ResNet-18}   & \textbf{83.16}        \\
    \textbf{GNA + margin loss+RL}      & \textbf{ResNet-18}   & \textbf{85.04}   \\
    \bottomrule
\end{tabular}}
\vspace{0pt}

\label{results:office}
\end{table}

%\begin{table}[h]
%  \caption{Clustering Hyper-Parameter Search for Domain Adaptation}
%  \label{table:layer-wise optimal number}
%  \begin{minipage}{0.5\textwidth}
%  \scalebox{0.65}{
%  \centering
%  \begin{tabular}{lllllllllllllllllllll}
%    \toprule
%    %\multicolumn{1}{2}{2}{c}{Accuracy(\%)} & Standard deviation                   \\
%    layer  &1&2&3&4&5&6&7&8&9&10&11&12&13&14&15&16&17&18&19&20   \\
%    
%    \midrule
%    # filters  &64&64&64&64&64&128&128&128&128&128&256&256&256&256&256&512&512&512&512&%512  \\
%    # clusters (manual)   &32&32&32&32&32&32&32&32&32&32&32&32&32&32&32&64&64&64&64&64 % \\
%    # clusters (greedy)   %&64&32&64&32&32&64&32&32&32&32&256&128&256&128&128&256&256&512&512&512 \\
%    # clusters (RL)       %&32&32&64&32&32&64&32&32&32&64&64&64&128&64&128&256&512&512&512&512 \\
%    \bottomrule
%  \end{tabular}}
%\end{minipage}
%\end{table}

\subsection{Transfer Learning}

In the domain adaptation experiment that has same classes, we showed our proposed method outperforms the baseline approaches. We can apply the grouping method to a task that the source classes are different from the target classes. We consider the task of transfer learning, where the $k$-shot learning task (target) is different from the pre-trained network task (source). 
%\todo{There was never really an explanation why we need to do transfer learning after domain adaptation. Is there is a qualitative difference? If yes, why is this evaluation approprate for evaluating our method?} 
Our pre-trained network is the ResNet-18 architecture trained with the classes on the CIFAR-100 dataset while the $k$-shot learning task is the classes on CIFAR-10 dataset. For transfer learning setting, we select the classes that different from ten target classes as source classes.
%\todo{This does not seem like transfer learning. They are pretty much the same dataset. Since you don't have time to do more experiments you'll need to explain why this is a transfer experiment even though they are really similar in nature. The classes are different right? Emphasize the classes and put less emphasis on the datasets in the writing}. 
Our action space for this experiment is the number of possible clusters in each layer $a = \{2^0,2^1,2^2,2^3,2^4\}$. We consider two different $k$-shot settings, one with $k=1$ and another with $k=10$. The $k$-shot data are chosen randomly from the target training set for fine-tuning and we evaluate on the entire target test set. The performance of our proposed approach is compared to the baselines in Table \ref{result:cifar} both for one-shot learning as well as for 10-shot learning. Our proposed margin loss improves the accuracies of the grouping method as well as fine-tuning. The accuracies of our grouping methods are higher than the fine-tuning result. Thus, the proposed method with RL search outperforms the baseline fine-tuning approach by 6.37\% in $1$-shot learning and 4.3\% in $10$-shot learning.

%\todo{Totally missing any analysis and discussion of results. I don't even know if the proposed method did better.}

\begin{table}[tb]
\caption{Experimental Evaluation: (Top) one-shot transfer learning from CIFAR-100 to CIFAR-10 and (Bottom) 10-shot transfer learning from CIFAR-100 to CIFAR-10.}
%  \begin{minipage}{0.2\textwidth}
\centering
  \scalebox{0.8}{
  \centering
  \begin{tabular}{ll}
    \toprule
    \cmidrule{1-2}
    Method & Accuracy(\%)\\
    \midrule
    % Late fusion & --      \\
    % Daume       & --      \\
    Fine-Tuning & 29.58 \\
    Fine-Tuning + margin loss & 33.44 \\
    \textbf{GNA} & \textbf{32.70}       \\
    \textbf{GNA+margin loss}         & \textbf{34.43}       \\
    \textbf{GNA+margin loss+greedy}         & \textbf{33.50}       \\
    \textbf{GNA+margin loss+RL}         & \textbf{35.95}       \\
    \bottomrule
  \end{tabular} 
  } \\(a) 1-shot learning\\
%\end{minipage}
%\quad
%\begin{minipage}{0.5\textwidth}
  \scalebox{0.8}{
  \centering
  \begin{tabular}{ll}
    \toprule
    \cmidrule{1-2}
    Method & Accuracy(\%)\\
    \midrule
    % Late fusion & --      \\
    % Daume       & --      \\
    Fine-Tuning & 56.00 \\
    Fine-Tuning + margin loss & 57.32 \\
    Fine-Tuning + triplet loss & 58.17 \\
    \textbf{GNA} & \textbf{57.96}       \\
    \textbf{GNA+margin loss}         & \textbf{59.05}       \\
    \textbf{GNA+triplet loss}         & \textbf{58.56}       \\
    \textbf{GNA+triplet loss+greedy}         & \textbf{58.56}       \\
    \textbf{GNA+triplet loss+RL}     & \textbf{60.30}       \\
    \bottomrule
  \end{tabular} 
  } \\(b) 10-shot learning\\
%\end{minipage}
\label{result:cifar}
\end{table}

\begin{figure}[tb]
    \centering
    \subfigure[one-shot]{\includegraphics[width=0.30\textwidth]{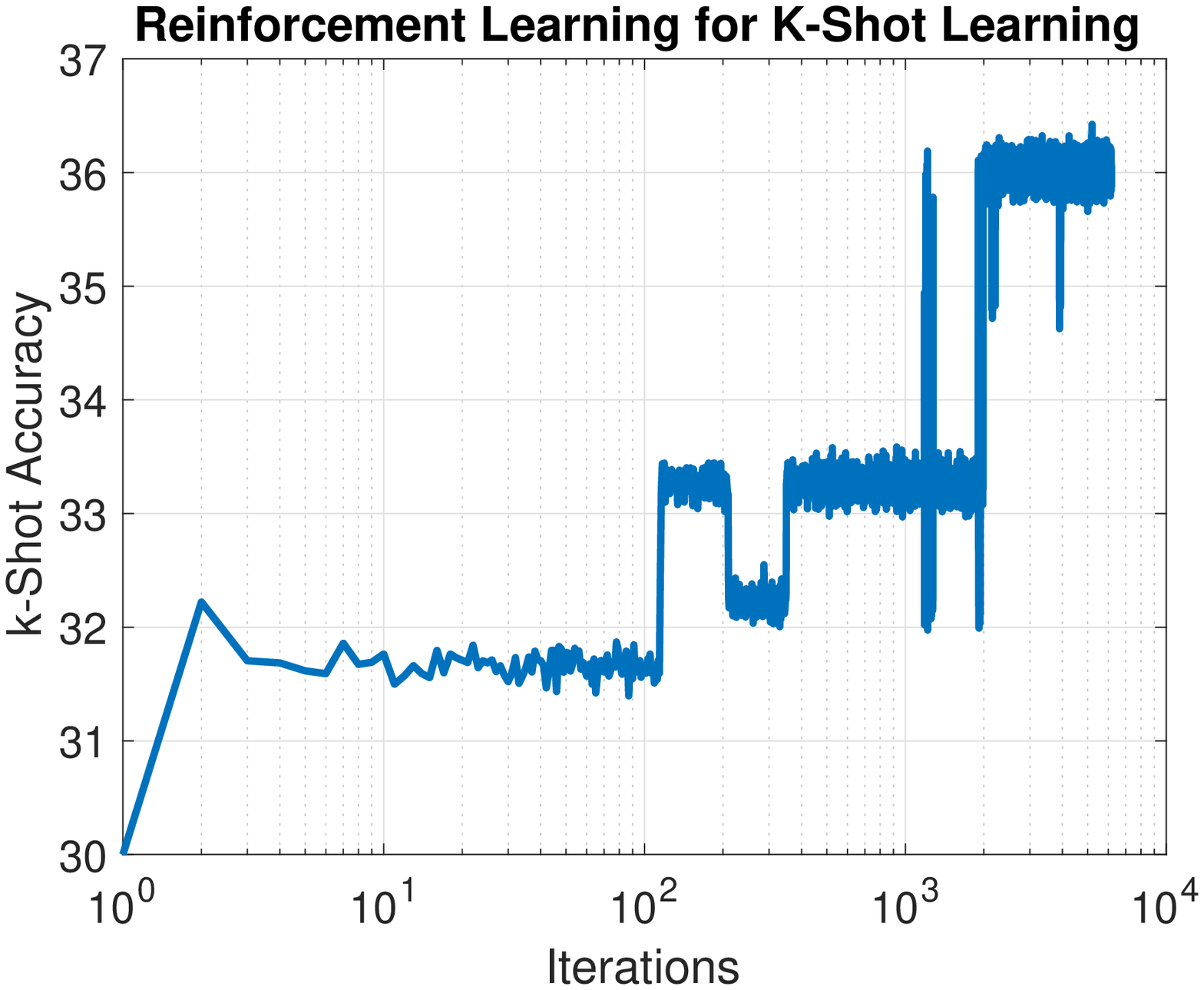}}
    \subfigure[10-shot]{\includegraphics[width=0.30\textwidth]{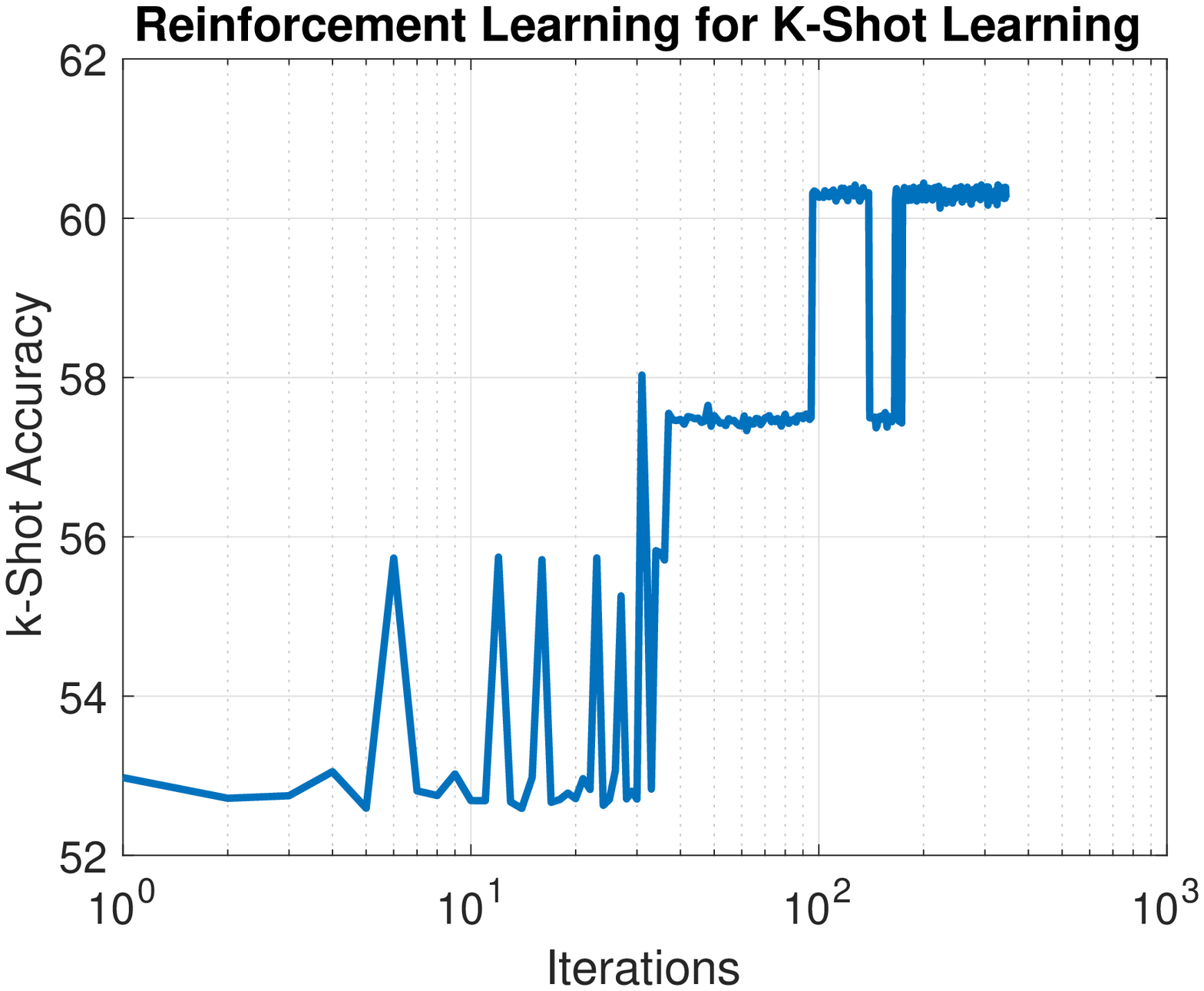}}
    \caption{Progression of hyper-parameter search through reinforcement learning through the iterations for $K$-shot transfer learning from CIFAR-100 to CIFAR-10}
    \label{fig::cifar}
\end{figure}

\subsection{Effect of Sample Size K} 

In this experiment we are interested in comparing the performance of our proposed approach as we vary the difficulty of the $k$-shot learning paradigm. We consider different ranges of $k$, the number of samples per category. Table \ref{table:k variation accuracy} presents the results of GNA with clustering and standard fine-tuning without clustering as we vary $k$. Unsurprisingly, the performance decreases and there is greater uncertainty as $k$ is lowered to one-shot learning. But we observe a consistent improvement in performance with our clustering approach in comparison to the standard fine-tuning procedure.
\begin{table}[h]
  \caption{$k$-shot classification performance as a function of number of samples per category}
  \label{table:k variation accuracy}
  \centering
  \begin{tabular}{lll}
    \toprule
    %\multicolumn{1}{2}{2}{c}{Accuracy(\%)} & Standard deviation                   \\
    The number  & {Accuracy(\%)}                    \\
    \cmidrule{2-3} 
    of \\clustering     & w/o clustering & w/ clustering \\
    \midrule
    25 shot  & 80.41  & 84.48 \\
    20 shot  & 76.90  & 82.09  \\
    15 shot  & 81.63  & 84.01 \\
    10 shot  & 70.64  & 72.88 \\
    5 shot   & 68.84  & 68.93 \\
    1 shot   & 52.25  & 53.77  \\
    \bottomrule
  \end{tabular}\\(a) Accuracy
  \\
  
  \begin{tabular}{lll}
    \toprule
    %\multicolumn{1}{2}{2}{c}{Accuracy(\%)} & Standard deviation                   \\
    The number  & Standard deviation                   \\
    \cmidrule{2-3} 
    of clustering     & w/o clustering & w/ clustering \\
    \midrule
    25 shot  & 5.68   & 0.76\\
    20 shot  & 2.66   & 1.44 \\
    15 shot  & 5.16   & 0.79\\
    10 shot  & 5.31   & 3.79\\
    5 shot   & 6.95   & 2.72\\
    1 shot   & 10.02  & 10.66 \\
    \bottomrule
  \end{tabular}\\(b) Standard deviation
\end{table}

\subsection{Effect of Clustering Across Layers}

It is commonly believed that most deep convolutional neural networks have highly redundant filters at the initial layers only. If this is indeed the case, applying our clustering method to layers other than the initial few layers should not be helpful. To test this hypothesis, we perform clustering to increasing number of layers, starting at the initial layers of the network. For this experiment we considered a pre-trained ResNet-18 network trained on a few categories in the CIFAR-10 dataset and used the other categories as the $k$-shot learning task. The results of GNA in Table \ref{table:OGM layer variation accuracy} surprisingly does not confirm our hypothesis. We found that all layers of the deep network did consist of redundant filters for the $k$-shot learning task. In fact, applying our method to all the layers of the network resulted in the best performance. This experiment suggests that large convolutional neural networks could potentially consist of redundant parameters even in the higher layers, necessitating search over this entire hyper-parameter space of parameter groupings. This motivates the need for efficient techniques to search the hyper-parameter space, like the one we proposed in this paper. %\todo{good. very strong writing. Make this the last experiment right before the conclusion to leave a better impression.}

\begin{table}[tb]
  \caption{$K$-shot classification performance as we vary layers where filters are clustered.}
  \label{table:OGM layer variation accuracy}
  \centering
  \begin{tabular}{lll}
    \toprule
    %\multicolumn{1}{2}{2}{c}{Accuracy(\%)} & Standard deviation                   \\
      & Accuracy(\%)             \\
    \cmidrule{2-3} 
    the number of layers & w/o clustering & w/ clustering    \\
    \midrule
    1 layer   & 80.41  & 82.87  \\
    3 layers  & 80.41  & 81.68  \\
    5 layers  & 80.41  & 82.98  \\
    7 layers  & 80.41  & 83.03  \\
    all       & 80.41  & 84.08  \\
    \bottomrule
  \end{tabular}\\(a) Accuracy
  
  \begin{tabular}{lll}
    \toprule
    %\multicolumn{1}{2}{2}{c}{Accuracy(\%)} & Standard deviation                   \\
      & Accuracy(\%)             \\
    \cmidrule{2-3} 
    the number of layers & w/o clustering & w/ clustering    \\
    \midrule
    1 layer   & 5.68 & 4.47   \\
    3 layers  & 5.68 & 4.01   \\
    5 layers  & 5.68 & 3.65   \\
    7 layers  & 5.68 & 3.11   \\
    all       & 5.68 & 0.76   \\
    \bottomrule
  \end{tabular}\\(b) Standard deviation

\end{table}

% \begin{table}[h]
%   \caption{Clustering Hyper-Parameter Search for Domain Adaptation}
%   \label{table:layer-wise optimal number}
%   \begin{minipage}{0.5\textwidth}
%   \scalebox{0.5}{
%   \centering
%   \begin{tabular}{lllllllllllllllllllllllllllll}
%     \toprule
%     %\multicolumn{1}{2}{2}{c}{Accuracy(\%)} & Standard deviation                   \\
%     layer &1&2&3&4&5&6&7&8&9&10&11&12&13&14&15&16&17&18&19&20&21&22&23&24&25&26&27&28   \\
    
%     \midrule
%     # filters  &16&160&160&160&160&160&160&160&160&160&320&320&320&320&320&320&320&320&320&640&640&640&640&640&640&640&640&640  \\
%     # clusters (manual)   &8&80&80&80&80&80&80&80&80&80&160&160&160&160&160&160&160&160&160&320&320&320&320&320&320&320&320&320  \\
%     # clusters (greedy)   &16&160&160&160&160&160&160&160&160&160&160&160&160&160&160&160&160&160&160&320&320&320&320&320&320&320&320&320 \\
%     # clusters (RL)       &16&80&80&80&80&80&80&80&80&  80& 320&  320&  320&  320&  320&  320&  320&  320&  320&  640&  640&  640&  640&  640&  640&  640&  640&  640 \\
%     \bottomrule
%   \end{tabular}}
% \end{minipage}
% \end{table}

\section{Conclusion}
\label{sec:conclusion}

In this paper we proposed a new regularization method for fine-tuning a pre-trained network for $k$-shot learning. The key idea of our approach was to effectively reduce the dimensionality of the network parameter space, by clustering the weights in each layer while ensuring intra-group similarity and inter-group orthogonality. To provide additional supervision to the $k$-shot learning problem we introduce a triplet loss to maximize the separation between the $k$-shot samples. Lastly, we introduced a reinforcement learning based approach to efficiently search over the hyper-parameters of our clustering approach. The experimental results  demonstrate that our proposed regularization technique can significantly improve the performance of fine-tuning based $k$-shot learning approaches.

%\newpage

%\clearpage

%\bibliographystyle{aaai}
%\bibliography{reference}

\end{document}